\documentclass[letterpaper, 10 pt, conference]{ieeeconf} 
\IEEEoverridecommandlockouts % This command is only needed if 
\usepackage[T1]{fontenc}
\usepackage{amsmath,amsfonts}
\usepackage{array}
\usepackage{cite}
\usepackage{stfloats}
\usepackage[caption=false,font=scriptsize]{subfig}
\usepackage{url}
\usepackage{graphicx}
\usepackage{paralist}
\usepackage{booktabs}
\usepackage[bookmarks=false]{hyperref}
\hypersetup{
colorlinks = true, % false: boxed links; true: colored links
hidelinks,
linkcolor=black, % color of internal links
citecolor=black, % color of links to bibliography
urlcolor=black % color of external links
}
\usepackage[capitalize]{cleveref}
\usepackage{tabularx}
\usepackage{siunitx}
\usepackage{verbatim}
\usepackage{blindtext}
\usepackage[inkscapelatex=false]{svg}

\usepackage[linesnumbered]{algorithm2e}
\RestyleAlgo{ruled}
\DontPrintSemicolon 

\usepackage{tikz}
\usetikzlibrary{arrows.meta}
\usetikzlibrary{arrows}
\usepackage{pgfplots}
\pgfplotsset{compat=1.16}

\usepackage{xcolor}
\definecolor{TUMBlue}{HTML}{0065BD}
\definecolor{DarkBlue}{HTML}{005293}

\definecolor{Orange}{HTML}{E37222}
\definecolor{Green}{HTML}{A2AD00}

\definecolor{Black}{HTML}{000000}
\definecolor{White}{HTML}{ffffff}

\definecolor{Gray}{HTML}{808080}
\definecolor{Graylight}{HTML}{CCCCCC}

\definecolor{OrangeCR}{HTML}{f1b514}
\definecolor{GreenCR}{HTML}{008000}

\newcommand{\frenetix}{Frenetix\xspace}

\newcommand{\lateraldisplacement}{\ensuremath{\mathit{d}}\xspace}
\usetikzlibrary{arrows, arrows.meta, external, matrix}

\newcommand{\coloredbox}[1]{
  \tikz{
    \draw[fill=#1] (0,0.0) rectangle (0.4,0.15);
  }
}

\newcommand{\goalareaCR}{
  \coloredbox{OrangeCR}
}

\newcommand{\arrowCR}{
  \tikz{
    \draw[{Circle[GreenCR,length=4pt]}-{triangle 45 [GreenCR,length=4pt]}, GreenCR] (0,0) -- (0.5,0);
    \draw[line width=2pt, GreenCR] (0.05,0) -- (0.3,0);
  }
}

\title{\LARGE \bf
MultiDrive: A Co-Simulation Framework\\ Bridging 2D and 3D Driving Simulation for AV Software Validation 
}

\author{Marc Kaufeld$^{1,*}$, Korbinian Moller$^{1,*}$, Alessio Gambi$^{2}$, Paolo Arcaini$^{3}$, Johannes Betz$^{1}$% 
\thanks{$^{1}$ The authors are with the Professorship of Autonomous Vehicle Systems, TUM School of Engineering and Design, Technical University of Munich, 85748 Garching, Germany; Munich Institute of Robotics and Machine Intelligence (MIRMI).}%
\thanks{$^{2}$A. Gambi is with the Center for Digital Safety \& Security at the Austrian Institute of Technology (AIT), 1210 Vienna, Austria.}%
\thanks{$^{3}$P. Arcaini is with the National Institute of Informatics (NII), Japan. He is supported by the ASPIRE grant No. JPMJAP2301, JST; and by Engineerable AI Techniques for Practical Applications of High-Quality Machine Learning-based Systems Project (Grant Number JPMJMI20B8), JST-Mirai.}%
\thanks{$^{*}$Shared first authorship.}
}

\begin{document}
\maketitle

%%%%%%%%%%%%%%%%%%%%%%%%%%%%%%%%%%%%%%%%%%%%%%%%%%%%%%%%%%%%%%%%%%%%%%%%%%%%%%%%
\begin{abstract}
%%%%%%%%%%%%%%%%%%%%%%%%%%%%%%%%%%%%%%%%%%%%%%%%%%%%%%%%%%%%%%%%%%%%%%%%%%%%%%%%
Scenario-based testing using simulations is a cornerstone of Autonomous Vehicles (AVs) software validation. 
So far, developers needed to choose between \emph{low-fidelity} 2D simulators to explore the scenario space efficiently, and 
\emph{high-fidelity} 3D simulators to study relevant scenarios in more detail, thus reducing testing costs while mitigating the sim-to-real gap.
This paper presents a novel framework that leverages multi-agent co-simulation and procedural scenario generation to support scenario-based testing across low- and high-fidelity simulators for the development of motion planning algorithms. %
Our framework limits the effort required to transition scenarios between simulators and automates experiment execution, trajectory analysis, and visualization.
Experiments with a reference motion planner show that our framework uncovers discrepancies between the planner's intended and actual behavior, thus exposing weaknesses in planning assumptions under more realistic conditions. 
Our framework is available at:
\url{https://github.com/TUM-AVS/MultiDrive}.
\end{abstract}

%%%%%%%%%%%%%%%%%%%%%%%%%%%%%%%%%%%%%%%%%%%%%%%%%%%%%%%%%%%%%%%%%%%%%%%%%%%%%%%%
%%% Keywords
%%%%%%%%%%%%%%%%%%%%%%%%%%%%%%%%%%%%%%%%%%%%%%%%%%%%%%%%%%%%%%%%%%%%%%%%%%%%%%%%
\vspace{0.2em}


%%%%%%%%%%%%%%%%%%%%%%%%%%%%%%%%%%%%%%%%%%%%%%%%%%%%%%%%%%%%%%%%%%%%%%%%%%%%%%%%
%%% Content
%%%%%%%%%%%%%%%%%%%%%%%%%%%%%%%%%%%%%%%%%%%%%%%%%%%%%%%%%%%%%%%%%%%%%%%%%%%%%%%%
%%%%%%%%%%%%%%%%%%%%%%%%%%%%%%%%%%%%%%%%%%%%%%%%%%%%%%%%%%%%%%%%%%%%%%%%%%%%%%%%
\section{Introduction}
\label{sec:introduction}

Autonomous vehicle (AV) technology rapidly progresses toward deployment in increasingly diverse operational design domains.
Consequently, general-purpose AVs must reliably handle a wide range of environments and traffic situations.
As discussed by Kalra and Paddock~\cite{kalra2016driving}, statistically demonstrating that AVs are as safe as human drivers would require testing AVs over several billion kilometers, making real-world validation economically infeasible~\cite{Riedmaier2020}.
Consequently, simulation-based development and scenario-based testing have become essential components of the AV software lifecycle.
For instance, motion planners, a core component of modular AV architectures~\cite{frenetix}, are often designed using low-fidelity simulators like CommonRoad (CR)~\cite{Althoff.2017} that provide fast feedback cycles and full controllability. 
Motion planning in low-fidelity testing environments is based on idealized assumptions, including perfect perception, simplified dynamics, and deterministic controller outputs.
However, due to sensor noise, imperfect actuation, and other discrepancies, design-time assumptions rarely hold in practice. 

Ignoring this ``sim-to-real gap'' may compromise safety and reliability in practice~\cite{StoccoTSE2023}. 
Therefore, developers must validate motion planners under conditions that more closely reflect reality.
Since real-world validation is too dangerous and expensive, developers could use high-fidelity simulators, such as BeamNG.tech~\cite{beamng_tech}, to validate their motion planners.
By implementing photorealistic and physically accurate simulations, these tools provide an appealing intermediate step to address the sim-to-real gap of low-level simulators.
However, running high-fidelity simulators is more time-consuming and requires developers to port scenario-based testing onto a different simulation, impacting testing efficiency and costs.

\begin{figure}[!t]
\centering
\subfloat[][High-fidelity 3D simulation in BeamNG.tech environment.]{\label{fig:multi-agent-sim-cr2}
\includegraphics[width=0.8\columnwidth]{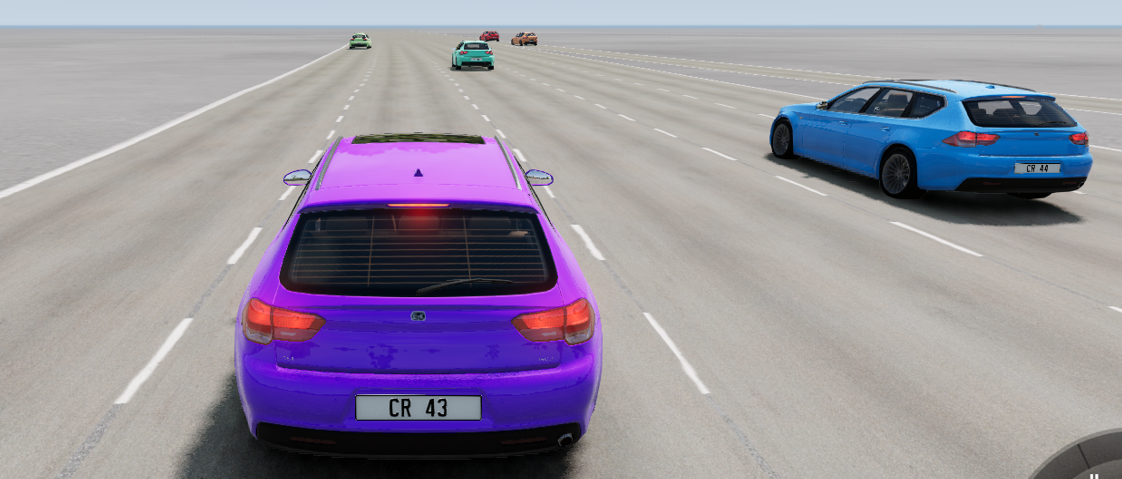}}
\\
\subfloat[][Low-fidelity 2D simulation in CommonRoad environment.]{\label{fig:multi-agent-sim-beamng1}
\includegraphics[width=0.8\columnwidth]{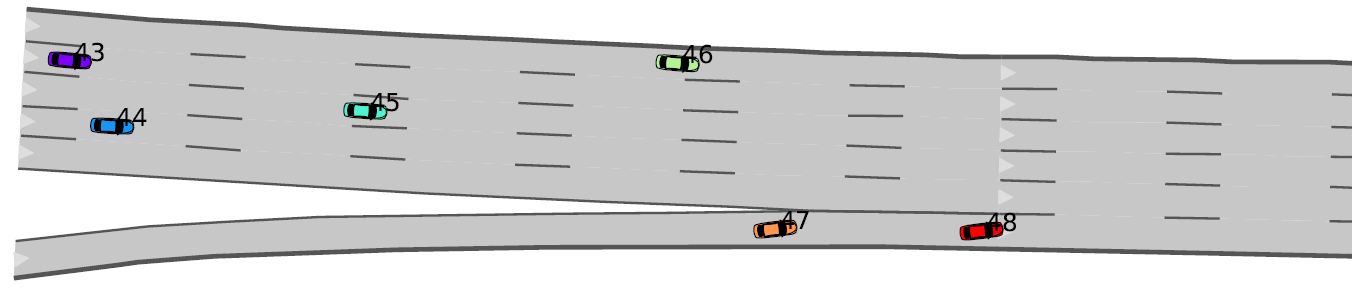}}
\caption{Visualization of a multi-agent scenario executed in both high-fidelity and low-fidelity simulators.}

\label{fig:intro}
\end{figure}

To address this problem, we devised a novel co-simulation framework which integrates low- and high-fidelity simulation platforms (see \cref{fig:intro}), and that supports procedural generation of multi-agent scenarios.
Our framework, built upon CommonRoad, \frenetix~\cite{frenetix, kaufeld2024}, and BeamNG.tech~\cite{beamng_tech}, allows the automatic generation of scenarios, their synchronized execution across both domains, and their analysis and visualization.
Therefore, it facilitates validating motion planners and analysing their robustness under realistic conditions.

\noindent The main contributions of this work are:
\begin{compactitem}
\item A novel co-simulation framework that enables the systematic validation of planning assumptions, revealing critical discrepancies in trajectory execution, particularly in interactive scenarios.
\item A fully automated, open-source pipeline supporting batch experiments, data collection, error analysis, and visualization.
\end{compactitem}

%%%%%%%%%%%%%%%%%%%%%%%%%%%%%%%%%%%%%%%%%%%%%%%%%%%%%%%%%%%%%%%%%%%%%%%%%%%%%%%%
\section{Related Work}
\label{sec:relatedwork}
Driving simulators and test scenarios generation are essential building blocks for the development of AV software, including motion planning algorithms~\cite{Tong.2020}. 
The next two sections summarize relevant existing work in these areas.

\subsection{Simulators for AV Software Testing}
Simulation frameworks vary in their purpose and level of fidelity, and generally fall into high- and low-fidelity simulators.
High-fidelity simulators are primarily used in end-to-end system validation where realistic vehicle dynamics and sensors matter~\cite{kaur2021}.
They can simulate vehicles using rigid-body (e.g., CARLA~\cite{Dosovitskiy.2017}, CarSim~\cite{carsim}, and LGSVL~\cite{rong2020}) and soft-body physics (e.g., BeamNG.tech~\cite{beamng_tech}) and offer advanced sensor simulation, scenario replay, and integration with external control stacks (e.g., via ROS bridges)~\cite{kaljavesi2024}. 
Low-fidelity simulators, such as CommonRoad~\cite{Althoff.2017} and NuPlan~\cite{Caesar.2021}, instead, are mostly employed for rapid development and validation of specific components, e.g., motion planning algorithms. They provide full controllability and deterministic executions. Thus providing fast feedback to developers and ease of scenario adjustment. However, they lack photorealism and can only provide simplified sensors.

To cope with these limitations, researchers integrated low-fidelity simulators with other tools.
For instance, Klischat et al.~\cite{Klischat.2019} integrated CommonRoad with the microscopic traffic simulator SUMO~\cite{Lopez.2018}, enabling model-based control of traffic participants via the Intelligent Driver Model (IDM)~\cite{Treiber.2017}. 
Kaufeld et al.~\cite{kaufeld2024} proposed a multi-agent extension to CommonRoad that allows testing of multiple interacting AVs.
Maierhofer et al.~\cite{Maierhofer2024} and Wuersching et al.~\cite{Wuersching2024}, instead, implemented interfaces connecting CommonRoad with CARLA~\cite{Maierhofer2024} and Autoware~\cite{Wuersching2024}. 
Like our framework, these interfaces allow motion planners implemented in CommonRoad to be tested in more realistic conditions.
However, unlike these interfaces, our framework can
\begin{inparaenum}[(1)]
\item automatically generate CommonRoad scenarios and instantiate them seamlessly into CommonRoad and BeamNG.tech, and
\item execute interactive scenarios that involve multiple motion planners. 
\end{inparaenum}
Additionally, while the Autoware bridge~\cite{Wuersching2024} only supports asynchronous execution, our framework enables synchronized simulations, allowing developers to test planning assumptions without interference from timing issues caused by computation and processing delays.

\subsection{Automatic Scenario Generation}
\label{ssec:sc}
Many approaches have been proposed to generate scenarios for automatically testing AV software~\cite{AVTestingSurvey}. For brevity, we report only some exemplary works below.
Interested readers may refer to a recent survey for a detailed overview~\cite{AVTestingSurvey}.
Most of the existing approaches focus on testing a single ego-vehicle and generate scenarios using combinatorial testing~\cite{cragSCICO2024,Stahl2019}, search algorithms~\cite{GambiISSTA2019,reqsViolADSsASE21,avoidCollICST2020}, or fuzzing~\cite{LiISSRE2020}.
More recent approaches integrate large language models into the scenario generation to promote the diversity of the generated scenarios~\cite{TangASE2024} and focus on testing interacting, non-cooperative AVs~\cite{DBLP:conf/icse/HuaiCANLWCG23,iv25-to-appear}.

%%%%%%%%%%%%%%%%%%%%%%%%%%%%%%%%%%%%%%%%%%%%%%%%%%%%%%%%%%%%%%%%%%%%%%%%%%%%%%%%
\section{Methodology}
\label{sec:methodology}

Our framework enables the systematic evaluation of motion planning methods for autonomous driving, across different levels of simulation fidelity. It is built around a shared scenario library accessible to both low- and high-fidelity simulators, ensuring that results from both are synchronized through a unified evaluation pipeline. This modular setup allows for iterative testing of planners across a wide range of conditions without the need of redundant reimplementation of scenarios or evaluation logic. Additionally, it enables developers to assess the robustness of the AV software by directly comparing its behavior across varying idealized and realistic execution conditions.

\begin{figure*}[!t]
\centering
\includegraphics[width=0.8\linewidth, trim=35 40 15 40, clip]{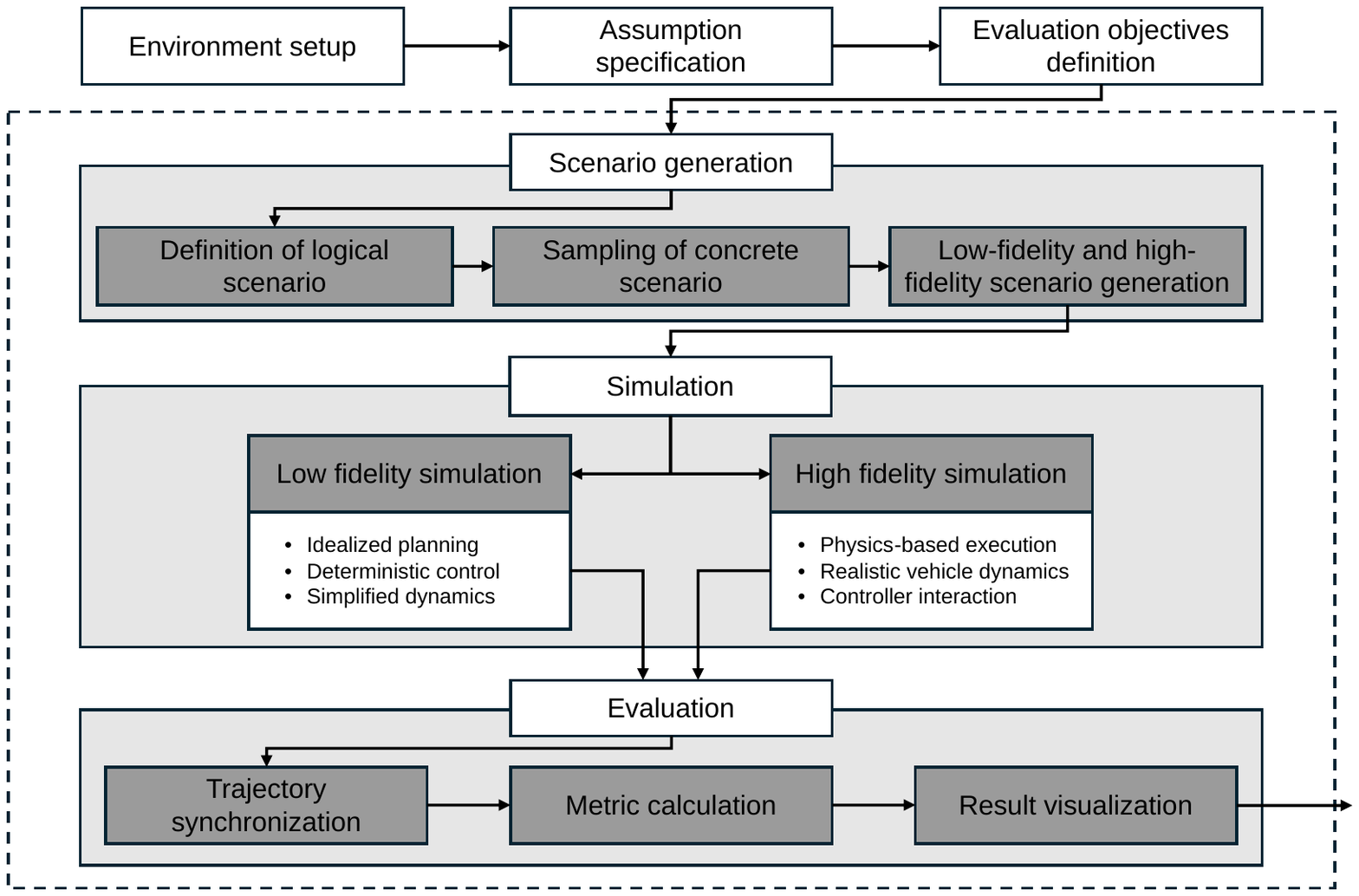}
\caption{Overview of the proposed co-simulation framework for validating AV software across low-fidelity and high-fidelity environments. The framework covers scenario generation, independent simulation at different fidelity levels, and synchronized evaluation through trajectory comparison and metric-based analysis.}
\label{fig:framework}
\end{figure*}

During the development of AV software, developers make various assumptions about the environment, sensors, vehicle dynamics, and other software modules. For example, they may test certain types of scenarios, rely on perfect environmental perception, use simplified vehicle dynamics (e.g., a unicycle model), or treat planned and executed trajectories as identical. 
In addition, they define specific evaluation metrics and testing procedures to assess the performance of the software.
These assumptions and metrics form the foundation for the testing and validation framework illustrated in \cref{fig:framework} and 
described in the following sections.

\subsection{Scenario Generation}
\label{subsec:ScenarioGeneration}
Scenario generation aims to identify key scenarios, such as critical situations, by exploring the scenario space defined by developers.
As shown in \cref{fig:framework}, the process begins by defining a logical scenario~\cite{MenzelIV2018}, which defines the parameter space in which the search variables, such as the initial state of the ego-vehicle, can be altered.
The goal is to explore this space systematically, creating different scenarios by adjusting the search variables.
Our framework can accommodate any existing scenario generation strategy (see \cref{ssec:sc}), but it also adds the flexibility to combine varying fidelities to guide the search process. 
For simplicity, we will adopt a grid search approach for demonstration purposes in our evaluation. Specifically, we will investigate different types of road geometries by changing the radius and turning angle of a turn (see \cref{sec:impactRoadFeatures}).

\subsection{Multi-fidelity Simulation Pipeline}
\label{subsec:SimulationPipeline}
Because of their underlying assumptions, trajectories are executed differently in low- and high-fidelity simulators.
In low-fidelity simulators, perfect control, exact state knowledge, and deterministic dynamics are assumed, resulting in a trajectory obtained simply by enforcing the planned vehicle states. Thus, planned and actual states always match perfectly.
In contrast, high-fidelity simulators model physical laws more realistically including realistic actuation behavior, disturbances, and environmental complexities; so planned and actual trajectories can differ.
During early development, developers typically rely on low-fidelity simulators for faster iteration. However, for practical deployment, it is necessary to test under more realistic conditions, which requires switching to high-fidelity simulations.
This transition is often challenging because different simulation platforms offer different configurations and interfaces. Without a unified framework, multi-fidelity testing becomes costly and error-prone, since manual reimplementation is necessary.
Our framework alleviates this issue as follows:

Given a 2D scenario in the low-fidelity simulation environment of CommonRoad~\cite{Althoff.2017}, our framework replicates the same scenario in the high-fidelity simulation of BeamNG.tech~\cite{beamng_tech} in 3D.
In CommonRoad, road lanelets are defined by polygons. We map these lanelets to BeamNG.tech's DecalRoads, which are defined by a centre line and a road-width attribute. Subsequently, the road material and lane markings are specified, such that they match the CommonRoad specifications (see \cref{fig:intro}). Finally, vehicles are spawned corresponding to the initial states of the vehicles in CommonRoad to simulate in high-fidelity.

To execute scenarios, we adopt the synchronous multi-agent simulation approach for CommonRoad proposed in~\cite{kaufeld2024}, which also allows us to analyse cooperative and competitive behaviors, as demonstrated in~\cite{kaufeld2024} and~\cite{iv25-to-appear}.
Here, a scenario involves multiple independent AVs, each equipped with full planning capabilities. At every time step, all vehicles plan new trajectories based on their local observations, execute one step along their planned trajectory, and update their internal states accordingly.
The planned trajectories are passed to the high-fidelity simulator, where an embedded controller fitted to each tested vehicle translates the trajectories into low-level control actions such as throttle and steering commands, to respect the physical dynamics of the vehicle. The (accurate) vehicle states from BeamNG.tech are then fed back into CommonRoad and used as initial conditions for the next planning cycle, enabling us to quantify the differences between the planned and the physically realized trajectories during the simulation, as explained next.

\subsection{Evaluation}
\label{subsec:EvaluationPipeline}
Simulated scenarios are analysed to improve and validate motion planning algorithms and to provide useful feedback for guiding automated scenario generation.
By leveraging the CommonRoad ecosystem~\cite{DBLP:conf/ivs/LinA23, kaufeld2024} and \frenetix~\cite{frenetix} as a reference motion planner, our framework can automatically compute several metrics assessing scenarios' criticality.

To further uncover issues related to the sim-to-real gap, we extend the evaluation pipeline by comparing low- and high-fidelity simulation results through alignment of simulated trajectories across different fidelity levels.
Following~\cite{Wuersching2024}, we compute error metrics that assess accuracy in position, orientation, and velocity of AVs.
Additionally, our platform allows for fast and intuitive visual inspection by rendering scenarios and error metrics. With this feedback, developers can quickly identify scenarios where the trajectory differences are minimal, indicating that the development assumptions align with reality, or where discrepancies are significant, suggesting that the AV software or the planning assumptions need revision.

%%%%%%%%%%%%%%%%%%%%%%%%%%%%%%%%%%%%%%%%%%%%%%%%%%%%%%%%%%%%%%%%%%%%%%%%%%%%%%%%

\section{Experimental Validation}
\label{sec:results}
We evaluated our framework's capabilities across different simulation fidelities, focusing on scenario generation consistency, assumption validation under varying conditions, and computational efficiency.

First, we verified that consistent scenarios can be generated and instantiated across low- and high-fidelity simulation environments.
Second, we demonstrated how our framework can identify planning assumption violations by systematically varying logical scenarios, road geometries, and vehicle models.
Finally, we evaluated the computational efficiency of simulations at different fidelity levels by comparing execution times.

For the evaluation, we used the \frenetix open-source motion planner~\cite{frenetix} as a test subject.
All experiments were conducted on a laptop with a 13th Gen Intel\textregistered~Core\texttrademark~i7-1365U @ 1.80~GHz, 32~GB of RAM, and running Windows 10 Enterprise Version 22H2. Simulations were executed within a WSL2 environment based on Ubuntu 22.04.5 LTS.

\subsection{Scenario Generation Consistency}

\begin{figure}[!t]
\centering
\subfloat[][High-fidelity simulation of an intersection scenario in BeamNG.tech.]{\label{fig:multi-agent-sim-beamng2}
\includegraphics[width=0.95\linewidth]{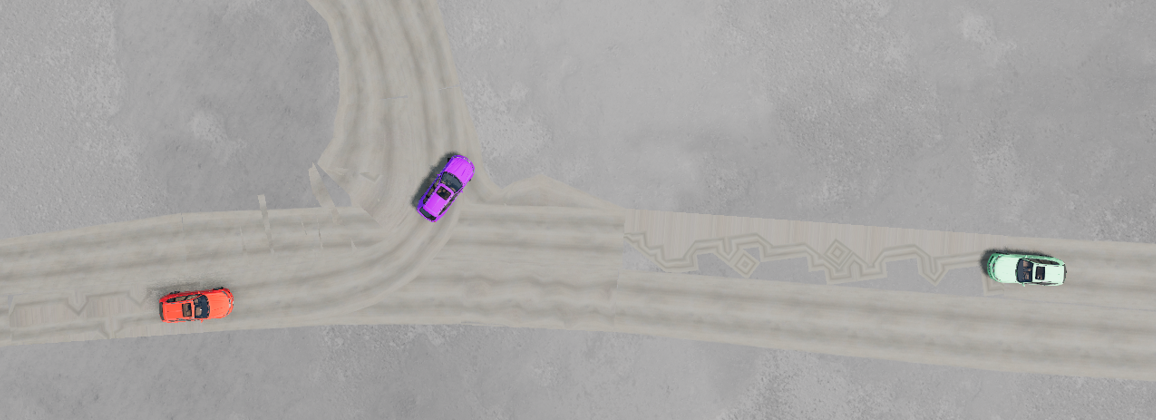}}\\
\subfloat[][Corresponding low-fidelity simulation in CommonRoad.]{\label{fig:multi-agent-sim-cr1}
\includegraphics[width=0.95\linewidth]{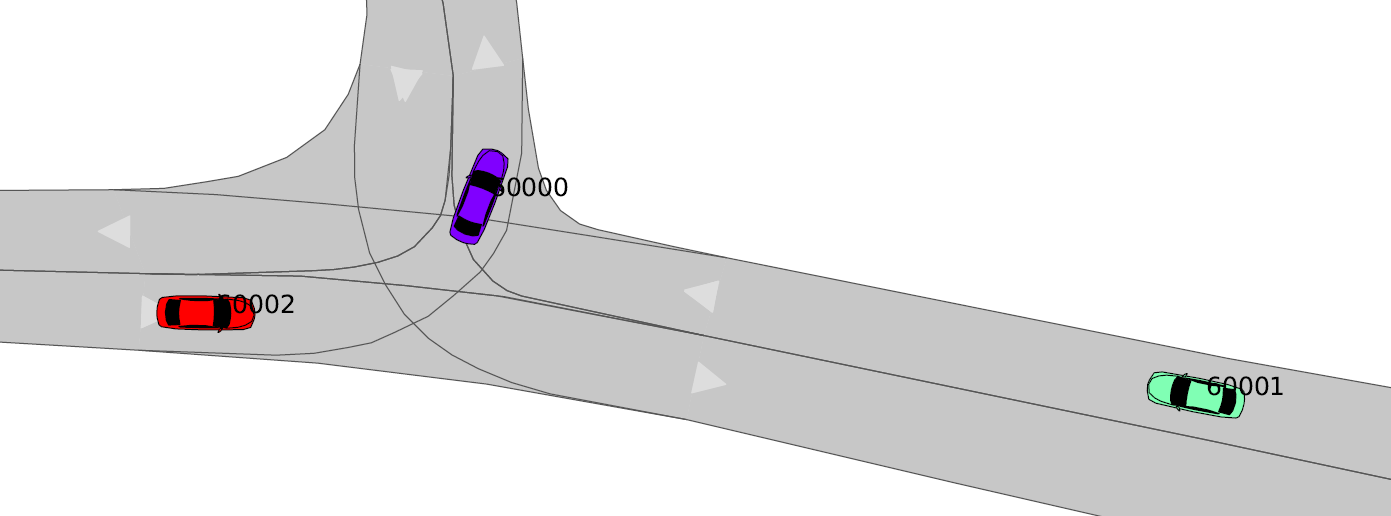}} 
\caption{Comparison of an intersection scenario executed in high-fidelity and low-fidelity simulators. Despite visual artefacts on the non-traveled lanelets in the high-fidelity rendering, the logical structure of the scenario and the interaction between vehicles are preserved across both environments.}

\label{fig:multi-agent-sim}
\end{figure}
\begin{figure*}[!b]
\centering
\subfloat[][$\gamma = \SI{0}{\degree}$]{\label{fig:angle0} \fbox{\includegraphics[height=0.5cm, trim={0.0cm 0.0cm 0.0cm 0.0cm},clip]{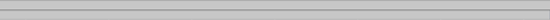}}} \\
\subfloat[][$\gamma = \SI{30}{\degree}$]{\label{fig:angle30} \fbox{\includegraphics[height=2.5cm, trim={0.0cm 0.0cm 0.0cm 0.0cm},clip]{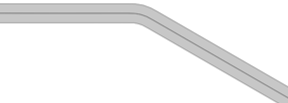}}} \hspace{0.5cm}
\subfloat[][$\gamma =\SI{60}{\degree}$]{\label{fig:angle60} \fbox{\includegraphics[height=2.5cm, trim={0.0cm 0.0cm 0.0cm 0.0cm},clip]{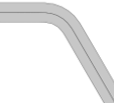}}} \hspace{0.5cm}
\subfloat[][$\gamma = \SI{90}{\degree}$]{\label{fig:angle90} \fbox{\includegraphics[height=2.5cm, trim={0.0cm 0.0cm 0.0cm 0.0cm},clip]{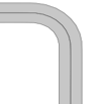}}} \hspace{0.5cm}
\subfloat[][$\gamma =\SI{120}{\degree}$]{\label{fig:angle120} \fbox{\includegraphics[height=2.5cm, trim={0.0cm 0.0cm 0.0cm 0.0cm},clip]{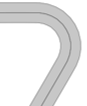}}}
\caption{Examples of generated scenarios with a curve radius of $r = \SI{10}{\meter}$ and varying curve angles \(\gamma\) from \SI{0}{\degree} to \SI{120}{\degree}. Only the relevant curve segments are shown for visualization. The full scenarios include longer road segments and define both a start and a goal position for the ego vehicle.}
\label{fig:curvature_examples}
\end{figure*}
To validate the scenario generation capabilities of our framework, we generated several scenarios in CommonRoad and instantiated them in BeamNG.tech. 
For each of them, we checked whether our framework could generate the corresponding BeamNG.tech simulation, and whether the vehicles' initial positions and orientations reported by BeamNG.tech matched those in CommonRoad. Additionally, we visually inspected the BeamNG.tech simulation to evaluate the rendering of the roads. Since our framework supports single- and multi-agent scenario simulations, we included both types of scenarios in our evaluation.
In total, we considered $78$ single-agent scenarios and $2$ multi-agent scenarios (see~\cref{fig:intro} and~\cref{fig:multi-agent-sim}).

In all cases, our framework could correctly instantiate scenarios in BeamNG.tech, such that the initial vehicles' position and orientation reported by BeamNG.tech perfectly matched those in CommonRoad.
Regarding the road rendering, for all scenarios except for the intersection one, we verified that road geometry, materials, and lane markings (when defined) in BeamNG.tech matched their CommonRoad counterparts.
For the intersection scenario (\cref{fig:multi-agent-sim}), we noticed a few visual artefacts that affected the road rendering. However, these artefacts were purely visual and did not impact the logical structure or execution of the scenario, which could still be successfully completed. 

\subsection{Evaluation of Planning Assumptions}

To investigate our framework's ability to uncover violations of planning assumptions, we conducted a systematic evaluation based on a grid search over different road geometries and using different vehicle models.

\subsubsection{Impact of Road Features}\label{sec:impactRoadFeatures}
We generated scenarios by varying the geometry of a turn (see~\cref{fig:curvature_examples}).
We defined turns with two parameters, radius and turning angle, and varied them to cover segments (i.e., left/right turns and straights) with increased complexity.

We simulated each scenario in low-and high-fidelity simulation and measured the lateral discrepancy \lateraldisplacement between the planned trajectories from both simulations.
Lateral displacement is a critical metric in trajectory following, especially in scenarios involving sharp turns or evasive maneuvers where even small tracking errors can lead to safety-critical events such as lane departures~\cite{Liu2023}.
We used BeamNG.tech's vehicle model (ETK 800), which better matches the model assumed by the motion planner (CommonRoad's Type~2~\cite{wuersching2021vehiclemodels}), approximating a BMW 320i.

\begin{figure}[!t]
\centering
\includegraphics[width=\linewidth]{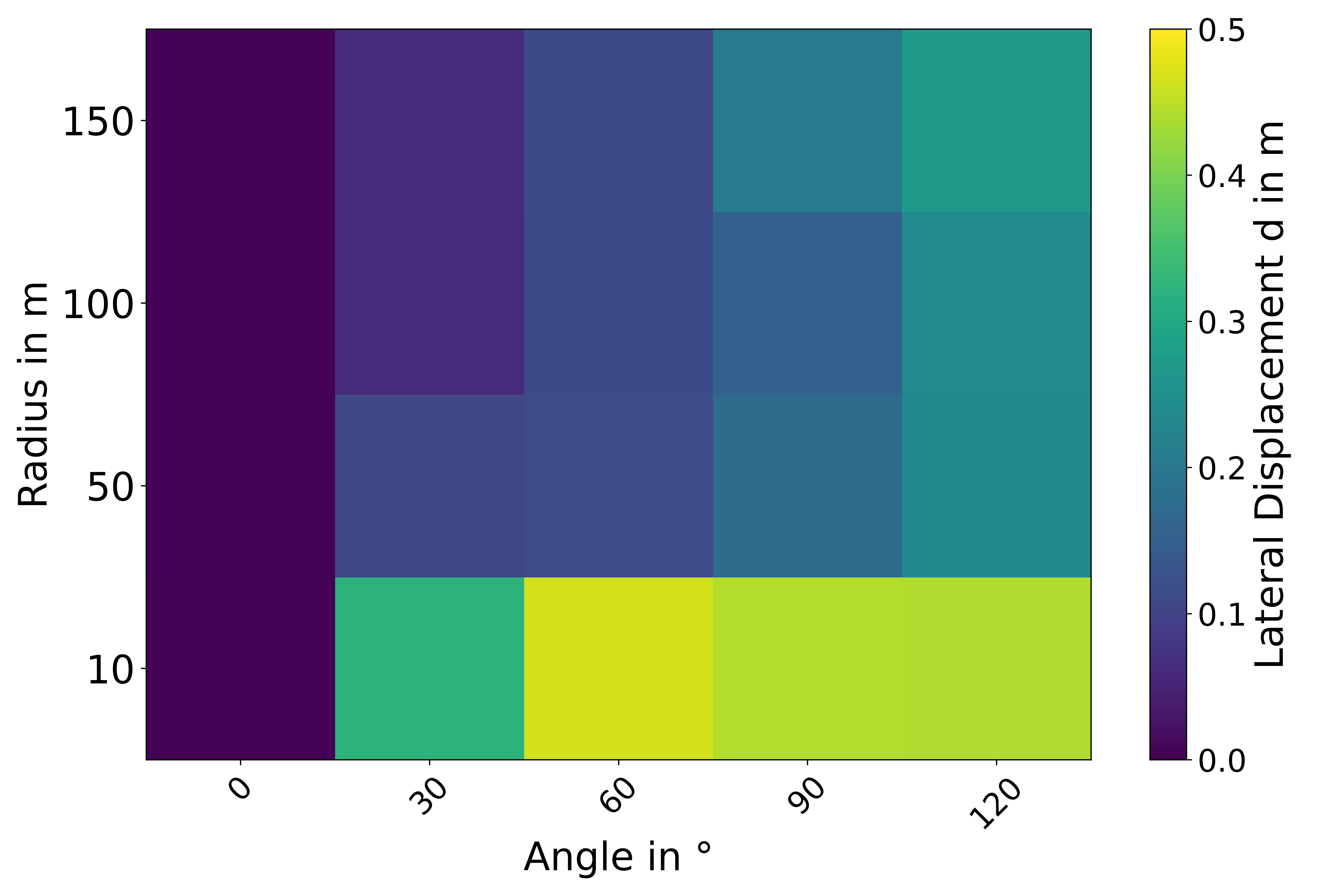}
\caption{Heatmap of the absolute lateral displacement error \(d\) between low-fidelity and high-fidelity simulations, measured across varying curve radii \(r\) and curve angles \(\gamma\). Brighter colors indicate larger deviations.}

\label{fig:curvature_heatmap}
\end{figure}

We report \lateraldisplacement across all tested combinations of curve radii and positive turning angles in~\cref{fig:curvature_heatmap}.
However, we do not report \lateraldisplacement for negative turning angles, as it follows a similar distribution due to the symmetry of the scenarios. 
From the figure, we observe that scenarios with smaller radii and larger turning angles, which correspond to sharper turns, exhibit increased lateral deviations, clearly indicating under which conditions the idealized assumptions used in \frenetix become less reliable.
Based on this observation, we conclude that our framework can be used to systematically evaluate planning assumptions and accelerate the identification of scenarios in which they are not met.

\subsubsection{Impact of Vehicle Model Disparities}
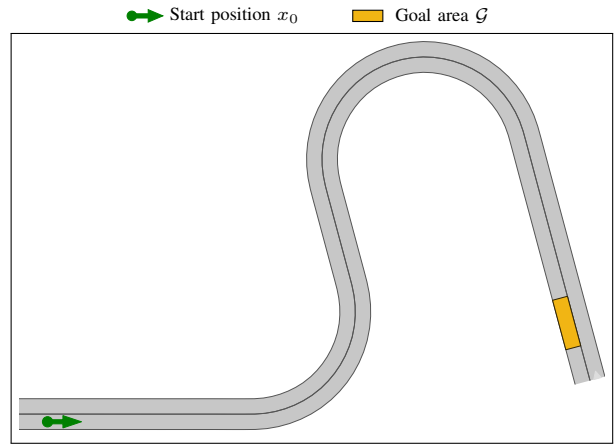
\begin{figure}[!t]
\centering
\begin{tikzpicture}[font=\scriptsize]
 % Start Position
 \node[inner sep=0pt] at (0.0,0) {\arrowCR};
 \node[align=left, anchor=west] at (0.2,0) {Start position $x_0$};

 % Goal Area
 \node[inner sep=0pt] at (3.0,0) {\goalareaCR};
 \node[align=left, anchor=west] at (3.2,0) {Goal area $\mathcal{G}$};
\end{tikzpicture}
\fbox{\includegraphics[width=0.9\linewidth, trim={5cm 1.5cm 3cm 2cm}, clip]{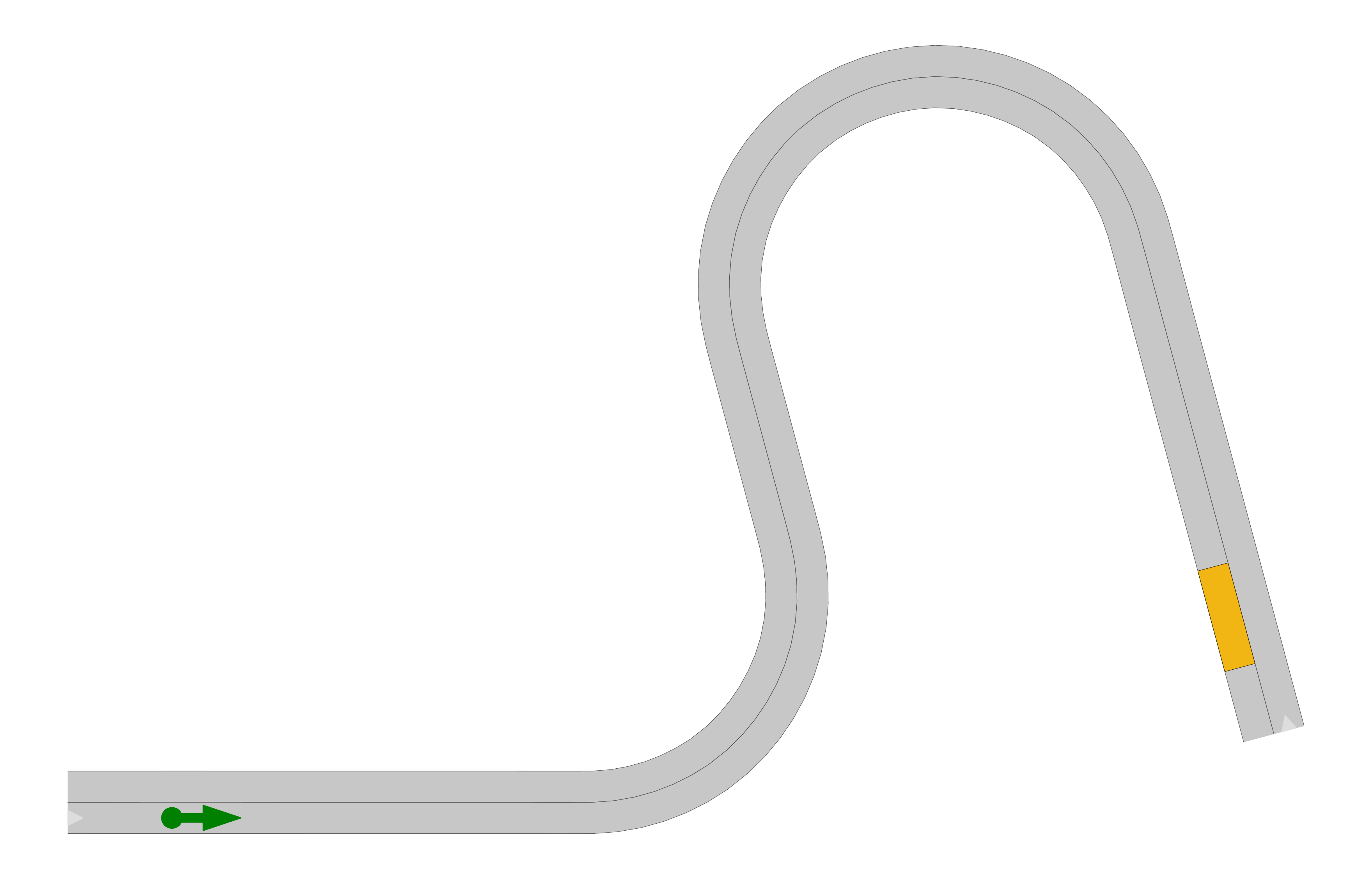}}
\caption{Example scenario used to assess the impact of mismatched vehicle models. The scenario includes a left-hand and right-hand curve as well as straight segments, enabling the evaluation of vehicle behavior under different driving conditions.}
\label{fig:vehicle_model_example}
\end{figure}

Besides generating scenarios with various road configurations, our framework can also simulate different vehicle models in high fidelity. This feature enables developers to assess the robustness of planning assumptions about vehicle dynamics by comparing the execution of scenarios with the same low-fidelity configuration but with varying vehicle dynamic models in BeamNG.tech. To illustrate the impact of mismatched vehicle dynamics, we selected a representative scenario involving a curved road segment (see~\cref{fig:vehicle_model_example}) 
and simulated it in low fidelity, using CommonRoad's Type~2 vehicle model, and in high fidelity using three different vehicle models available in BeamNG.tech: 
ETK 800, Hopper, and Piccolina.\footnote{The complete list of BeamNG.tech's vehicle models is available at: \url{ https://documentation.beamng.com/official_content/vehicles/}}
We selected these three models because they represent touring cars, off-road vehicles, and small city cars, i.e., vehicles with different structural properties and varying driving characteristics.
We compared the motion behavior resulting from different vehicle models and computed two key metrics: the lateral displacement \lateraldisplacement as a function of the traveled distance \(s\), and the velocity profile \(v\) as a function of time \(t\).
The lateral shift 
\lateraldisplacement provides insights into the ability of a controller to track the intended path. In contrast, deviations in the velocity profile \(v\) can reveal differences in acceleration and deceleration behavior that could affect maneuver feasibility and safety margins.
We report the results for lateral displacement and velocity profiles in~\cref{fig:lateral_displacement_vehicle_models} and \cref{fig:velocity_profiles_vehicle_models}, respectively.
\begin{figure*}[!t]
\centering
\begin{minipage}{0.48\textwidth}
\centering
\input{figures/d_over_s.tex}
\caption{Lateral displacement \(d\) over traveled distance \(s\) for different vehicle models.}
\label{fig:lateral_displacement_vehicle_models}
\end{minipage}
\hfill
\begin{minipage}{0.48\textwidth}
\centering
\input{figures/v_over_t.tex}
\caption{Velocity profile \(v\) over time \(t\) for different vehicle models.}
\label{fig:velocity_profiles_vehicle_models} 
\end{minipage}
\end{figure*}
Given the reference planner \frenetix and its planning assumptions, we observe that the shape of the road has a more significant impact on lateral displacement than the vehicle model used in the high-fidelity simulation. All three high-fidelity simulations exhibit similar trends, with the small city car showing the greatest deviation.
In contrast, the velocity is tracked accurately in the multi-fidelity simulation by the embedded controller, but again, the small car exhibits the largest offset.

The controller is able to follow the planned velocity quite accurately, but the small car also has the highest offset.
This highlights that our framework enables the evaluation of discrepancies caused by mismatched vehicle models between low- and high-fidelity simulations. This capability provides a systematic approach to uncover planning assumptions tied to specific vehicle dynamics. 
Such evaluations are crucial for developing more robust, vehicle-independent planning strategies. Additionally, these insights can guide the adaptation of planners to specific vehicle models used in control modules when deploying on real-world vehicles.
\subsection{Computational Efficiency Analysis}
Comparing the time to execute the simulation in low- and high-fidelity environments lets developers better balance the trade-off between fast but inaccurate simulations and slow but accurate ones. 
For this purpose, we measured the time required to execute a set of ($78$) scenarios using CommonRoad and BeamNG.tech.
We report basic statistics about the execution time of low- and high-fidelity simulations in~\cref{tab:runtime_comparison}.
\begin{table}[!hbt]
\centering
\caption{Comparison of simulation runtimes across 78 scenarios for low- and high-fidelity simulation.}
\label{tab:runtime_comparison}
\begin{tabular}{lcc}
\toprule
Statistic & Low-Fidelity & High-Fidelity \\
\midrule
Minimum & \SI{12.01}{\second} & \SI{122.12}{\second} \\
Maximum & \SI{19.12}{\second} & \SI{206.88}{\second} \\
Mean & \SI{14.11}{\second} & \SI{182.07}{\second} \\
Median & \SI{13.78}{\second} & \SI{183.67}{\second} \\
Std. Deviation & \SI{1.48}{\second} & \SI{10.98}{\second} \\
\bottomrule
\end{tabular}
\end{table}
The table shows that low-fidelity simulation is significantly (about $10$ times) faster than high-fidelity simulation, confirming our expectations and strengthening our belief that combining low- and high-fidelity simulations offers important advantages when developing and validating AV software.

% %%%%%%%%%%%%%%%%%%%%%%%%%%%%%%%%%%%%%%%%%%%%%%%%%%%%%%%%%%%%%%%%%%%%%%%%%%%%%%%%
\section{Discussion}
\label{sec:discussion}
Our results demonstrate how our framework combines low- and high-fidelity simulations to validate motion planners and their underlying planning assumptions.
By varying road geometries, vehicle models, and configurations through automated scenario generation, developers can identify scenarios where planning assumptions, such as assumed vehicle dynamics, may not hold when testing under more realistic conditions in high-fidelity simulations. 

In our exemplary study, we use a grid search to explore the parameter space of the road geometry. However, the different fidelity levels could be used in combination to guide advanced search methods toward discovering violations of developers' assumptions, for example, by identifying scenarios that pass low-fidelity simulators but fail in high-fidelity ones. 
This opens up new possibilities for an efficient exploration of the scenario space.

A current drawback of the proposed framework is that the high-fidelity simulator cannot enforce arbitrary vehicle states. Thus, currently only scenarios with vehicles starting from a standing state can be tested. Future improvements will tackle the problem of instantiating scenarios with arbitrary initial vehicle states.

\section{Conclusion and Outlook}\label{sec:conclusion}
This paper introduced a novel co-simulation framework for validating AV software in 2D and 3D driving simulators.
Our framework combines CommonRoad and BeamNG.tech through a shared scenario library and an automated evaluation pipeline, enabling developers to validate AV software under both idealized and physically realistic conditions. 
As exemplified by its application on an open-source motion planning algorithm, our framework supports the targeted identification of violated planning assumptions, supports multi-agent scenarios, and provides quantitative metrics for assessing execution deviations.
In contrast to existing interfaces binding low- and high-level simulators, our framework supports automated scenario generation and procedural content generation. Therefore, it explicitly addresses the testing and validation of planning assumptions through scenario variation and cross-simulation comparison.

Future work includes extending the evaluation to a broader set of planning algorithms, scenarios, and configurations, and integrating advanced scenario generation approaches to systematically study the relationship between planner design choices and real-world performance indicators. Additionally, more advanced procedural content generation approaches to enhance both the accuracy and the visual quality of the scenario rendering will be investigated.

\section*{Acknowledgment}

Marc Kaufeld and Korbinian Moller contributed equally to this work and share first authorship. Together, they developed the overall concept of the framework, designed the evaluation strategy, conducted the experimental validation, and played a major role in shaping the methodology and structure of the research. Alessio Gambi was responsible for the co-simulation implementation and the development of the automated evaluation pipeline. He also actively contributed to the overall conceptualization of the framework. Paolo Arcaini led the scenario generation activities, critically evaluated and refined the underlying ideas of the paper, and supported the manuscript review and proofreading process. Johannes Betz critically reviewed the article, approved the final version for publication, and supported the research project. The authors would like to thank the Munich Institute of Robotics and Machine Intelligence (MIRMI) for their continuous support.

%%%%%%%%%%%%%%%%%%%%%%%%%%%%%%%%%%%%%%%%%%%%%%%%%%%%%%%%%%%%%%%%%%%%%%%%%%%%%%%%

%%%%%%%%%%%%%%%%%%%%%%%%%%%%%%%%%%%%%%%%%%%%%%%%%%%%%%%%%%%%%%%%%%%%%%%%%%%%%%%%
%%% Bibliography
%%%%%%%%%%%%%%%%%%%%%%%%%%%%%%%%%%%%%%%%%%%%%%%%%%%%%%%%%%%%%%%%%%%%%%%%%%%%%%%%
\bibliographystyle{IEEEtran}
\bibliography{literatur}
\end{document}